\documentclass[runningheads]{llncs}
\usepackage[T1]{fontenc}
\usepackage{graphicx,verbatim}
\usepackage{booktabs, tabularx, xcolor}
\usepackage{booktabs}   
\usepackage{multirow}   
\usepackage{array}      
\usepackage{amsmath}
\usepackage{hyperref}
\begin{document}
\title{IRIS: An Intelligent Vision-Language System for Ocular Surface Diseases via Topic Tree and Scene-Driven VQA Generation}
\titlerunning{IRIS: An Intelligent Vision-Language System}
%
\author{Hao Wei\inst{1}\thanks{Equal contribution.}\orcidID{0000-0002-5719-8826} \and
Wenjin Qi\inst{2}\protect\footnotemark[1]\orcidID{0009-0004-9816-1423} \and
Dasen Dai\inst{1}\orcidID{0009-0000-3227-3140} \and Minqing Zhang\inst{1}\orcidID{0000-0002-7214-0569} \and Wu Yuan\inst{1}\orcidID{0000-0001-9405-519X}\thanks{Corresponding author (wyuan@cuhk.edu.hk).}}
\authorrunning{H. Wei et al.}
%
\institute{The Chinese University of Hong Kong, Shatin, Hong Kong, China \and
Westlake University, Xihu District, Hangzhou, Zhejiang Province, China}


  
\maketitle              
\begin{abstract}

While Large Vision-Language Models (VLMs) demonstrate remarkable generic capabilities, their clinical reasoning in specialized domains like ocular surface diseases (OSDs) is severely hindered by a paucity of high-fidelity, multimodal instruction-tuning data. To dismantle this data bottleneck, we introduce IRIS, an Intelligent Recognition and Interaction System tailored for fine-grained OSD understanding via external eye photography. First, we curate IRIS-120K, the largest and most comprehensive OSD visual question-answering (VQA) dataset to date. Crucially, to overcome the semantic shallowness of conventional image-caption pairs, we propose a synergistic data generation paradigm to explicitly inject clinical priors. Our data engine operates via a dual-branch framework: 1) a Topic Finding Tree (TFT) that hierarchically anchors visual features to precise anatomical and pathological concepts, enforcing rigorous medical deduction logic; and 2) a Scene-driven strategy that synthesizes role-adaptive clinical dialogues to ensure pragmatic generalization. By explicitly aligning a compact 4B-parameter VLM on this structurally enriched corpus, IRIS achieves state-of-the-art performance, comprehensively outperforming both generalist and specialized medical VLMs with up to 34B parameters. Our findings underscore that structured knowledge injection profoundly prevails over sheer parameter scaling, unlocking the potential for resource-efficient, expert-level AI deployment on mobile edge devices for scalable OSD screening. Code, datasets, and model weights will be publicly released by this \href{https://github.com/hwei-hw/IRIS}{repo}.


\end{abstract}
\section{Introduction}


Ocular surface diseases (OSDs), including dry eye disease (DED) and infectious keratitis, represent a formidable global health challenge and the fifth leading cause of blindness~\cite{maehara2025importance,yonehara2025use}. DED afflicts up to 34\% of adults, while infectious keratitis is a leading cause of blindness, which is largely preventable if diagnosed early~\cite{graham2024machine,maehara2025importance}. Despite this prevalence, expert-level diagnosis traditionally mandates specialized stationary equipment (e.g., slit-lamp biomicroscopes)~\cite{lu2025artificial,yonehara2025use}, exacerbating healthcare disparities in resource-limited settings. Consequently, the ubiquity of high-resolution smartphone cameras has catalyzed mobile eye health (MeHealth) into a transformative frontier, offering an unprecedented opportunity for accessible, large-scale OSD screening via external eye photography~\cite{wang2025leveraging}.


Ophthalmic AI has rapidly evolved from task-specific classifiers (e.g., CorneAI ~\cite{maehara2025artificial,maehara2025importance}) towards multimodal foundation models like RETFound~\cite{zhou2023foundation}, VisionFM~\cite{qiu2023visionfm} and EyeCLIP~\cite{shi2025multimodal}. While these architectures demonstrate robust generalization for retinal conditions, a critical modality disparity persists: current models exhibit a profound bias towards the posterior segment. For instance, within the 2.77-million-image EyeCLIP dataset, external eye photographs account for a negligible 0.5\%. Similarly, EyecareGPT~\cite{li2025eyecaregpt} predominantly targets fundus and OCT imagery. This leaves a severe data and reasoning vacuum in the fine-grained understanding of external eye pathologies.


While generic Large Vision-Language Models (VLMs) show potential for interactive diagnostics~\cite{srinivasan2025can,yeh2024insight}, recent benchmarking exposes their stark clinical limitations in ophthalmology, often producing hallucinated interpretations lacking precise anatomical grounding~\cite{yeh2024insight,li2025large}. Even domain-specific medical VLMs (e.g., HuatuoGPT-Vision~\cite{HuatuoGPT-Vision}, Lingshu~\cite{Lingshu}) are inherently bottlenecked by the flat, unstructured nature of PubMed image-caption pairs. Consequently, they consistently fail to spatially anchor morphological features to specific anatomical structures (e.g., distinguishing a corneal infiltrate from an adjacent iris nodule)---a rigorous spatial deduction step critical for accurate OSD diagnosis~\cite{jin2025systematic}.


To bridge this diagnostic gap, we introduce \textbf{IRIS} (\textbf{I}ntelligent \textbf{R}ecognition \& \textbf{I}nteraction \textbf{S}ystem for OSDs). We curate \textbf{IRIS-120K}, the largest comprehensive OSD visual question-answering (VQA) dataset to date. To explicitly overcome the semantic misalignment of generic VLMs~\cite{jin2025systematic,yeh2024insight}, we engineer a Clinically-driven Data Engine composed of two novel paradigms:
1) \textbf{Topic Finding Tree (TFT):} A structured framework that hierarchically anchors visual features to 10 predefined ocular regions and pathological findings, forcing the model to internalize rigorous, step-by-step clinical deduction.
2) \textbf{Scene-Driven Generation:} A role-adaptive paradigm simulating clinical interactions across three primary users (Doctor, Patient, Student) and 12 scenarios, ensuring pragmatic utility in real-world MeHealth deployments.


By fine-tuning a highly compact 4B-parameter VLM on this enriched corpus, IRIS achieves state-of-the-art (SOTA) performance, overwhelmingly surpassing generalist and medical models possessing up to 34B parameters. Our findings underscore a vital paradigm shift: meticulous structural knowledge injection and explicit anatomical grounding---rather than mere parameter scaling---are definitive keys to enabling highly efficient, expert-level AI deployment on mobile edge devices for scalable OSD screening~\cite{jin2025systematic,wang2025leveraging}.

The primary contributions of this work are summarized as follows:

\begin{itemize}
    \item \textbf{A Dedicated OSD Interactive Corpus:} We curate and release IRIS-120K, successfully filling a major void in high-fidelity, multimodal instruction-tuning data specifically tailored for external eye photography.
    \item \textbf{Knowledge-Injected Data Engines:} We pioneer the TFT and Scene-Driven strategies to explicitly model spatial ocular anatomy and clinical personas, fundamentally resolving the interpretability and pragmatism deficits inherent in existing medical VLMs.
    \item \textbf{Highly Efficient Foundation Model:} We open-source IRIS-4B, a lightweight SOTA clinical VLM optimized for resource-constrained environments. It demonstrates unprecedented parameter efficiency and fine-grained visual-textual grounding compared to massively scaled baselines.
\end{itemize}



\section{Methodology}

\begin{figure}[t]
    \centering
    \includegraphics[width=\textwidth]{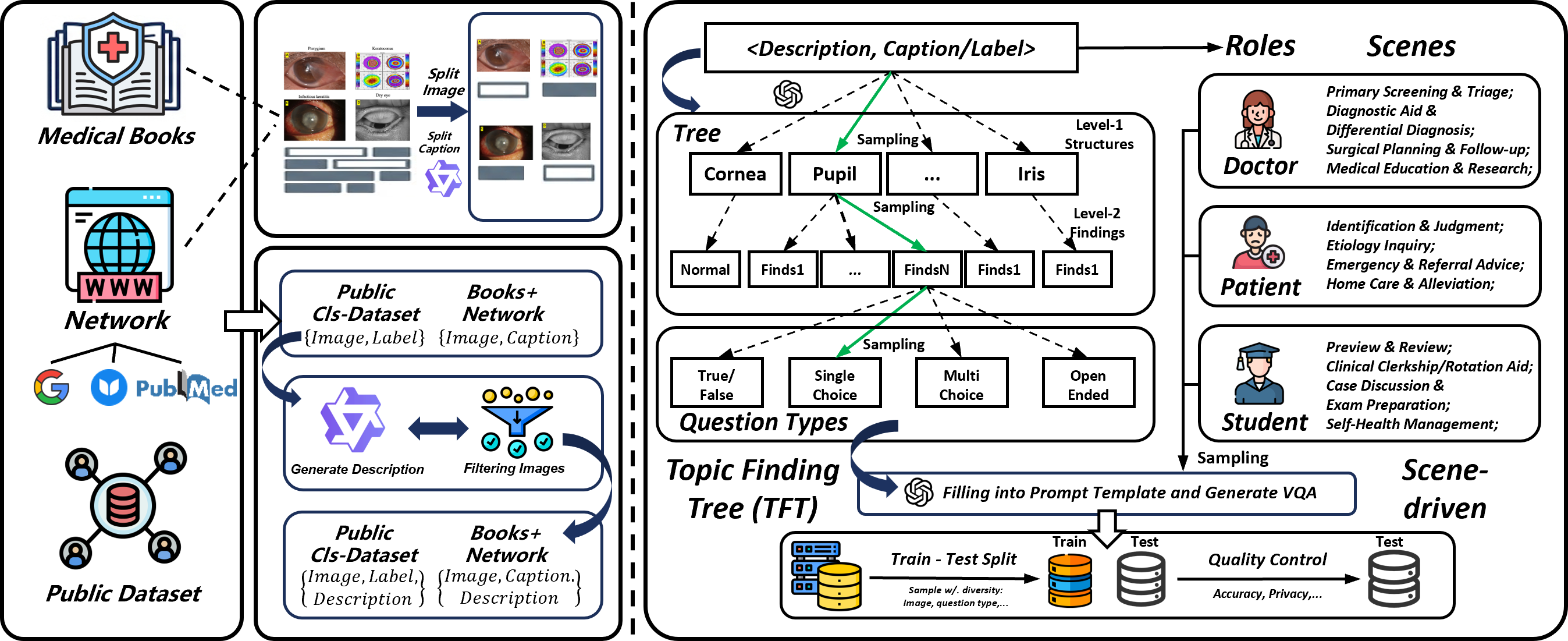} 
    \caption{Overview of our Clinically-driven Data Engine. The pipeline aggregates heterogeneous ocular images, and utilizes a dual-branch VQA generation paradigm: Topic Finding Tree (TFT) for rigorous clinical reasoning and Scene-driven generation for role-adaptive interactions.}
    \label{fig:framework}
\end{figure}

To bridge the gap between the poor zero-shot performance of existing VLMs and the clinical demand for OSD screening, we propose a novel Clinically-driven Data Engine (Fig.~\ref{fig:framework}). Rather than simply scaling up model parameters, our framework curates \textbf{IRIS-120K}, the largest external eye VQA dataset to date. By innovating Topic Finding Tree (TFT) and Scene-driven generation paradigms, we inject rigorous medical reasoning and role-adaptive empathy into lightweight models, enabling highly efficient edge deployment on patient smartphones.

\subsection{Large-Scale Data Curation and Preprocessing}
\label{subsec:data_collection}

To construct our Data Engine, we aggregate external eye images into \textit{image-caption} or \textit{image-label} pairs from three heterogeneous sources: \textbf{Medical Books} (3.1K cases), \textbf{Network Data} (93.5K samples from WeChat, PubMed, and Google Lens), and \textbf{Public Datasets} (35.4K). Note that these counts are the results before any filtering. To resolve domain noise and format inconsistencies, we design source-specific preprocessing pipelines.

\noindent\textbf{Books and PubMed Literature.} 
Medical literature contains high-value but complex multi-panel figures that severely degrade VLM image-text alignment. Sourcing from textbooks and BIOMEDICA~\cite{biomedica} (a biomedical image-caption archive), we deploy a rigorous three-stage pipeline~\cite{open-pmc}. First, Qwen2.5-VL-7B~\cite{Qwen2.5-VL} filters non-external eye images and categorizes data into Paper-Single (standalone) and Paper-Multi (compound). For \textit{Paper-Multi}, a YOLOv7-based detector~\cite{pathasst} segments sub-figures, and Qwen2.5-VL-72B~\cite{Qwen2.5-VL} uses these bounding boxes to semantically map the overarching monolithic caption to specific sub-figure crops. A secondary VLM filter eliminates irrelevant content (e.g., fundus) exposed after splitting. Captions are further enriched using main-text references. Book data undergoes an identical fine-grained alignment.

\noindent\textbf{Web Resources.} 
To capture real-world long-tail diversity, we use curated book images as visual seeds for Google Lens reverse searches, extracting alt-texts as captions. Concurrently, we scrape ophthalmic WeChat articles, which natively provide structured \textit{(image, disease name, symptom description)} triplets. Both sources undergo strict VLM-based filtering to guarantee external-eye relevance.

\noindent\textbf{Public Datasets Adaptation.} 
We integrate high-quality classification and segmentation datasets. To adapt segmentation data, we introduce an area-based heuristic: if over 50\% of the images contain lesion bounding boxes occupying $\ge$50\% of the total image area (meaning localized pathology visually dominates), the entire dataset is repurposed for classification. All data is standardized as \textit{(image, label)} pairs.

\noindent\textbf{Quality-Aware Stratification.} 
Given varying raw source fidelity, we conduct sampling-based human evaluations on resolution, image-text alignment, and semantic richness to derive a strict quality hierarchy: \textbf{WeChat $>$ Books $>$ Paper-Single $>$ Public Cls $>$ Paper-Multi $>$ Web (Google Lens)}. This empirically established prior fundamentally directs our ``Quality-Aware Dynamic Sampling'' strategy (Sec.~\ref{subsec:vqa_generation}), ensuring high-fidelity data dominates the training distribution of our lightweight models.

\subsection{Knowledge-Driven VQA Generation \& Quality-Aware Sampling}
\label{subsec:vqa_generation}

Traditional ``image-caption'' pairs generate flat representations, failing to impart the ``anatomy-to-sign'' deduction logic required in clinical practice. To overcome this, we transform the curated images into interpretable VQA pairs using a dual-branch strategy. To enrich the semantic context and prevent LLM hallucinations, we first employ Qwen3-VL-32B~\cite{Qwen3-vl} to generate a detailed visual description $T_{desc}$ for each image. Combined with the original caption/label $T_{cap}$, this forms a rich textual surrogate $x = (T_{cap}, T_{desc})$, serving as the foundational semantic anchor for downstream generation.

\noindent\textbf{Topic Finding Tree (TFT) Generation.}
To force the model to ``think step-by-step like an ophthalmologist,'' we introduce TFT, a structured anatomical decomposition framework. The TFT hierarchically parses $x$ into a depth-2 logical tree: the first level comprises 10 predefined ocular anatomical regions $a_i \in \mathcal{A}$ ( \textit{Pupil, Cornea, Iris, Sclera-Conjunctiva Region, Eyelashes, Upper Eyelid, Lower Eyelid, Superior Palpebral Conjunctiva, Inferior Palpebral Conjunctiva, and Others }), and the second level extracts region-specific clinical findings $f \in F_{a_i}$. This explicitly establishes anatomical-pathological paths $\mathcal{T} = \{ (a_i, f) \mid a_i \in \mathcal{A}, f \in F_{a_i} \}$.

For each instantiated path, an LLM generates diverse question formats (e.g., True/False, Single-choice, Multiple-choice or Open-ended). Crucially, we prompt the LLM to construct a structured \textbf{4-stage reasoning chain} ($Think$): \textit{(1) Visual Observation} (identifying morphological features); \textit{(2) Clinical Correlation}; \textit{(3) Logical Deduction} (excluding differential diagnoses); and \textit{(4) Conclusion}. This explicit mapping ensures the generated VQA tuples $(Q, A, Think)$ are strictly grounded in localized clinical evidence.

\noindent \textbf{Scene-Driven Adaptation.}
To accommodate the diverse interaction needs of real-world edge deployments, we simulate conversational contexts across three primary user roles: $\mathcal{R} = \{\text{Patient, Doctor, Student}\}$. For each role, we define tailored clinical scenes $\mathcal{S}_r$ (e.g., triage advice for patients, differential diagnosis for doctors, as shown in the right panel of the Fig.~\ref{fig:framework}). Using scene-specific templates, the LLM generates interactions incorporating a sequential clinical chain-of-thought—\textit{Query Analysis, Evidence Extraction, Clinical Reasoning, Response Formulation}—ensuring the response tone and jargon density closely align the target user's medical literacy.

\noindent\textbf{Quality-Aware Dynamic Sampling \& Dataset Finalization.}
Exhaustive generation across all TFT nodes and clinical scenes would yield over 2 million highly redundant VQA pairs. This contradicts our fundamental objective of training a highly efficient, small-parameter model (2B/4B) capable of offline edge deployment. Therefore, we implement a quality-aware dynamic sampling protocol. Instead of uniform generation, we probabilistically assign combination strategies based on the predefined data quality hierarchy. High-quality sources trigger denser generation (e.g., sampling multiple TFT regions and scenarios simultaneously), while lower-quality web data is sparsely sampled to maximize information yield without redundancy.
For each image, 1--3 prompts are generated using either the TFT or Scene-Driven method.

\begin{figure}[t]
    \centering
    \includegraphics[width=\textwidth]{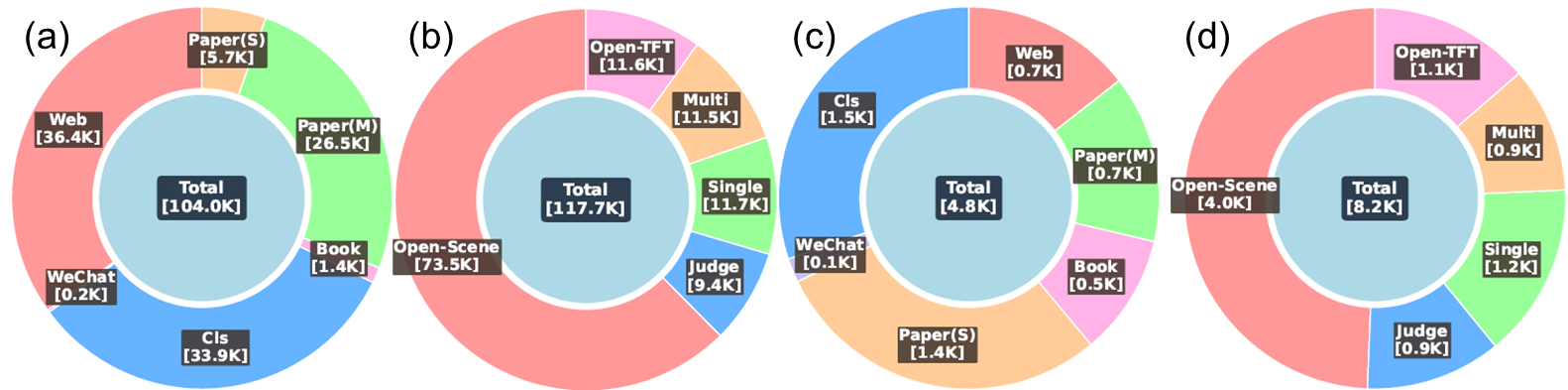} 
    \caption{Dataset statistics of IRIS-120K. (a)/(c) Distribution of image sources across Train/Test sets; (b)/(d) Distribution of generated VQA types (TFT vs. Scene-driven) in train and test, respectively.}
    \label{fig:stat}
\end{figure}

This balanced protocol generates approximately 130K initial VQA pairs. 
We apply pHash~\cite{phash} for strict image deduplication to prevent data leakage. An initial test pool of 9K VQA pairs undergoes a rigorous final quality control pass using 
GPT-5-mini to filter out factual hallucinations and privacy anomalies. This ultimately yields the \textbf{IRIS-120K}, comprising \textbf{117.7K} training pairs and an \textbf{8.2K} gold-standard test set (detailed statistics in Fig.~\ref{fig:stat}).

\begin{table*}[t]
\centering
\caption{Performance comparison of General (Gen), Medical (Med), and Our models across all five question types. \textit{Gen. Avg.} denotes the average score across the six open-ended generative metrics (\textbf{B}LEU and \textbf{R}OUGE). Best results are in \textbf{bold}.}
\resizebox{\textwidth}{!}{
\begin{tabular}{llccc|ccccccc}
\toprule
\multirow{2}{*}{Group} & \multirow{2}{*}{Model} & \multirow{2}{*}{Judge} & \multirow{2}{*}{Single} & \multirow{2}{*}{Multi} & \multicolumn{3}{c}{Tree-VQA} & \multicolumn{3}{c}{Scene-VQA} & \multirow{2}{*}{Gen.Avg.} \\
\cmidrule(lr){6-8} \cmidrule(lr){9-11}
 &  &  &  &  & B-1 & R-1-F & R-L-F & B-1 & R-1-F & R-L-F &  \\
\midrule
\multirow{7}{*}{Gen} 
 & GPT-5-mini & 69.52 & 72.08 & 67.48 & 16.48 & 11.31 & 21.81 & 13.73 & 22.02 & 11.81 & 16.19 \\
 & Internvl3.5-14B & 77.35 & 75.62 & 54.59 & 11.82 & 19.80 & 11.59 & 18.50 & 20.22 & 11.34 & 15.55 \\
 & Internvl3.5-8B & 72.59 & 75.37 & 49.41 & 10.52 & 17.94 & 10.64 & 15.42 & 18.68 & 10.17 & 13.90 \\
 & Qwen3-vl-32B & 77.57 & 79.98 & 62.64 & 13.65 & 20.79 & 13.06 & 19.22 & 20.36 & 11.84 & 16.49 \\
 & Qwen3-vl-8B & 69.52 & 73.06 & 57.41 & 7.77 & 16.30 & 8.88 & 13.33 & 17.09 & 9.18 & 12.09 \\
 & Qwen3-vl-4B & 62.43 & 70.68 & 51.98 & 6.19 & 14.24 & 7.83 & 10.86 & 16.39 & 8.63 & 10.69 \\
 & Qwen3-vl-2B & 69.31 & 62.36 & 44.15 & 4.82 & 15.88 & 9.06 & 7.15 & 16.59 & 8.96 & 10.41 \\
\midrule
\multirow{9}{*}{Med} 
 & HGPT-V-7B & 32.70 & 25.62 & 8.45 & 21.34 & 22.65 & 15.02 & 19.58 & 18.53 & 11.59 & 18.12 \\
 & HGPT-V-34B & 66.56 & 68.29 & 50.36 & 28.60 & 27.18 & 18.79 & 24.40 & 22.02 & 13.89 & 22.48 \\
 & HuluMed-7B & 83.07 & 49.09 & 26.61 & 31.91 & 30.17 & 20.58 & 26.12 & 25.49 & 15.84 & 25.02 \\
 & HuluMed-14B & 83.92 & 59.59 & 25.63 & 32.48 & 30.68 & 20.80 & 25.98 & 25.70 & 15.70 & 25.22 \\
 & HuluMed-32B & 83.70 & 60.05 & 30.39 & 33.02 & 31.06 & 21.11 & 26.02 & 25.92 & 15.90 & 25.51 \\
 & Lingshu-7B & 84.97 & 64.25 & 37.70 & 21.63 & 25.88 & 19.33 & 13.59 & 20.64 & 13.46 & 19.09 \\
 & Lingshu-32B & 84.76 & 84.43 & 59.71 & 28.04 & 29.85 & 21.92 & 18.08 & 23.66 & 15.13 & 22.78 \\
 & MG-4B & 66.88 & 60.38 & 36.22 & 16.13 & 22.68 & 16.52 & 17.76 & 19.94 & 12.27 & 17.55 \\
 & MG1.5-4B & 71.01 & 58.24 & 31.14 & 11.52 & 17.19 & 9.98 & 11.72 & 16.36 & 8.26 & 12.51 \\
\midrule
\multirow{5}{*}{Ours} 
 & IRIS-8B & 97.04 & 98.19 & 88.55 & 42.27 & \textbf{39.68} & \textbf{29.64} & 44.19 & \textbf{36.32} & \textbf{24.74} & \textbf{36.14} \\
 & \textbf{IRIS-4B} & \textbf{97.25} & \textbf{98.52} & \textbf{88.96} & \textbf{42.36} & 39.29 & 29.03 & \textbf{44.23} & 36.08 & 24.45 & 35.91 \\
 & IRIS-4B-TFT & 96.51 & 98.19 & 88.26 & 41.86 & 39.18 & 28.95 & 40.46 & 28.86 & 18.17 & 32.91 \\
 & IRIS-4B-Scene & 93.12 & 83.20 & 48.42 & 37.02 & 34.31 & 24.48 & 44.13 & 36.01 & 24.41 & 33.39 \\
 & IRIS-2B & 96.19 & 97.03 & 83.98 & 40.40 & 38.06 & 27.77 & 30.28 & 26.70 & 16.86 & 30.01 \\
\bottomrule
\end{tabular}
}
\label{tab:exp_results}
\end{table*}

\section{Experiments}

\noindent\textbf{Experimental Setup.} 
First, we We fine-tune Qwen3-VL via Low-Rank Adaptation on four A6000 GPUs to get IRIS series models, more details can be found our repo. To comprehensively evaluate multi-modal clinical capabilities, we benchmark IRIS against 16 SOTA VLMs, encompassing General-domain (GPT-5-mini~\cite{gpt5mini}, InternVL-3.5 series~\cite{Internvl}, Qwen3-VL series~\cite{Qwen3-vl}) and Medical-domain models (HuatuoGPT-Vision~\cite{HuatuoGPT-Vision}, HuluMed~\cite{HuluMed}, Lingshu~\cite{Lingshu}, Med-Gemma ~\cite{medgemma}). We employ Accuracy for closed-ended objective tasks (Judge, Single/Multi-choice) and standard generative metrics (BLEU-1~\cite{bleu}, ROUGE-1/L-F~\cite{rouge}) for open-ended interactions (Tree/Scene-VQA).



\noindent\textbf{Main Results.} 
Quantitative evaluations are summarized in Table~\ref{tab:exp_results}. The IRIS series establishes a new SOTA across all question types. Our optimal model, \textbf{IRIS-4B}, achieves near-perfect objective reasoning accuracies (97.25 on Judge, 98.52 on Single-choice) and dominates complex generative tasks (e.g., 42.36 on Tree-VQA B-1). Specifically, across the five primary metrics (Accuracies and B-1 scores), IRIS-4B attains an outstanding overall average of \textbf{74.26}.

Crucially, IRIS exhibits profound parameter efficiency, achieving ``cross-scale'' suppression. IRIS-4B significantly outperforms baselines up to 8$\times$ larger, outperforming the strongest medical VLM (Lingshu-32B, overall avg. 55.00) by 19.26 absolute points, and the best generic VLM (Qwen3-VL-32B, overall avg. 50.61) by 23.65 points. Furthermore, even our minimalist IRIS-2B (avg. 69.57) consistently beats all 32B/34B competitors. This validates that our specialized clinical data engine bridges the domain gap far more effectively than mere parameter scaling.



\noindent\textbf{Ablation Studies.} 
\textbf{1) Model Scaling:} Evaluating the architecture across 2B, 4B, and 8B scales reveals that clinical capability peaks at 4B. Scaling further to 8B introduces a slight performance plateau, indicating that 4B is the optimal capacity sweet-spot for learning our specialized representations without overfitting, ideal for edge deployment. 
\textbf{2) Data Composition:} We ablated the training components of IRIS-4B. Training exclusively on structured TFT data (\textit{IRIS-4B-TFT}) preserves strong logical deduction but degrades Scene-VQA generation. Conversely, relying solely on role-playing Scene data (\textit{IRIS-4B-Scene}) precipitates a substantial degradation in objective precision (e.g., Multi-choice accuracy plummets from 88.96 to 48.42). This confirms the essential synergy of our framework: TFT builds the rigorous medical logic, while Scene-adaptation provides the clinical interaction interface.

\begin{figure}[t]
    \centering
    \includegraphics[width=\textwidth]{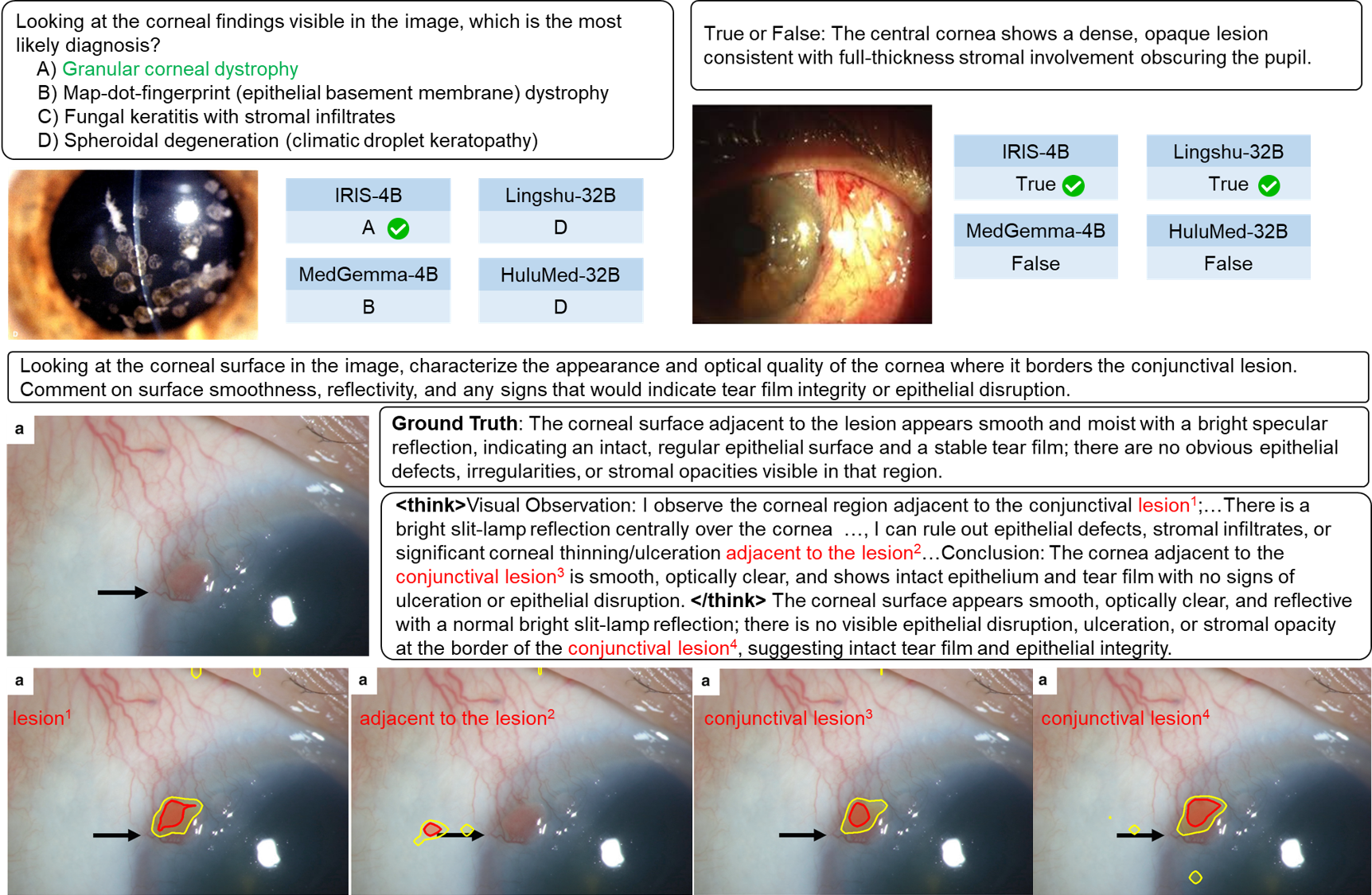} 
    \caption{Qualitative evaluation of IRIS-4B. \textbf{Top:} Accurate predictions on complex objective tasks, outperforming substantially larger medical baselines. \textbf{Bottom:} Demonstration of transparent reasoning and visual grounding. IRIS-4B articulates its logic via a \texttt{<think>} block and visually anchors generated clinical terms (red text) to localized pathological regions (attention contours: $>25\%$ yellow, $>50\%$ red).}
    \label{fig:example}
\end{figure}

\noindent\textbf{Qualitative Analysis \& Interpretability.} 
Fig.~\ref{fig:example} qualitatively validates the clinical reliability and transparency of IRIS. In complex closed-ended tasks (Fig.~\ref{fig:example}, top), IRIS-4B accurately diagnoses \textit{Granular corneal dystrophy} and verifies stromal involvement, overcoming the diagnostic inconsistencies and hallucinations that plague much larger medical baselines (e.g., Lingshu-32B). 

Crucially, for open-ended interactions (Fig.~\ref{fig:example}, bottom), IRIS-4B transcends ``black-box'' answering. Driven by our structured TFT training paradigm, it generates an explicit \texttt{<think>} block to articulate a rigorous clinical chain-of-thought---detailing visual observations and logical deductions---prior to formulating its final conclusion. Furthermore, IRIS-4B exhibits robust \textit{visual grounding}. Generated clinical terms (e.g., ``lesion'', ``adjacent to the lesion'') are spatially anchored to the input image via attention maps (delineated by $>25\%$ yellow and $>50\%$ red importance contours). While functioning conceptually akin to Class Activation Mapping (CAM) rather than exact pixel-level segmentation, this precise spatial-textual correlation confirms that IRIS-4B explicitly anchors its reasoning to authentic morphological cues rather than language priors, offering vital interpretability for reliable clinical triage.

\section{Conclusion}

In this paper, we introduced IRIS, a highly efficient and interpretable Vision-Language Model tailored for OSD screening. To overcome the severe domain gap in existing VLMs, we curated IRIS-120K, the largest external eye VQA dataset to date. By pioneering a Clinically-driven Data Engine that integrates the rigorous anatomical logic of a Topic Finding Tree with role-adaptive Scene-driven interactions, we successfully forced the model to internalize precise clinico-pathological deduction rather than superficial language priors.

Extensive evaluations demonstrate that our optimal model, IRIS-4B, establishes a new SOTA across diverse multi-modal questions, overwhelmingly outperforming general and medical baselines up to eight times its size. This empirical breakthrough highlights a vital insight: \textit{high-quality, structurally curated domain data profoundly prevails over sheer parameter scaling}. Furthermore, IRIS mitigates the ``black-box'' skepticism inherent in VLMs by seamlessly coupling an explicit step-by-step reasoning chain (\texttt{<think>}) with attention-based visual grounding, mirroring the transparent diagnostic workflow of human ophthalmologists. Ultimately, by achieving unprecedented accuracy and interpretability with minimal computational overhead, IRIS not only sets a robust benchmark for specialized medical VLMs but also clears the hardware barriers for privacy-preserving, completely offline edge deployment on patient smartphones, paving the way for accessible, large-scale epidemiological screening.

\bibliographystyle{splncs04}
\bibliography{mybibliography}
%




\end{document}